\documentclass[runningheads]{llncs}
\usepackage[T1]{fontenc}
\usepackage{graphicx} % Required for inserting images
\usepackage{booktabs}   % for \toprule etc.
\usepackage{tabularx}   % auto-width columns (X)
\usepackage{makecell}   % \makecell
\usepackage{array}      % column modifiers
\usepackage{float}      % [H] specifier (optional; you can use [htbp] instead)
\usepackage{amsmath}    % math
\usepackage{threeparttable}
\usepackage{bm}
\begin{document}
\title{Synthetic Data–Guided Feature  Selection for Robust Activity Recognition in Older Adults}
\author{Shuhao Que\inst{1}\orcidID{0000-0003-2806-5255} \and
Dieuwke van Dartel\inst{1,2}\orcidID{0000-0002-3556-4522} \and
Ilse Heeringa\inst{1} \and
Han Hegeman\inst{1,2} \and
Miriam Vollenbroek-Hutten\inst{1,3}\orcidID{0000-0001-8730-1487} \and
Ying Wang\inst{1}\orcidID{0000-0003-4385-9117}
}
\authorrunning{S. Que et al.}
\institute{University of Twente, Drienerlolaan 5, 7522 NB Enschede, the Netherlands \and
Ziekenhuis Groep Twente, Zilvermeeuw 1, 7609 PP Almelo, the Netherlands \and
Medisch Spectrum Twente,  Koningstraat 1, 7512 KZ Enschede, the Netherlands
}
\maketitle              % typeset the header of the contribution
\begin{abstract}
Physical activity during hip fracture rehabilitation is essential for mitigating long-term functional decline in geriatric patients. However, it is rarely quantified in clinical practice. Existing continuous monitoring systems with commercially available wearable activity trackers are typically developed in middle-aged adults and therefore perform unreliably in older adults with slower and more variable gait patterns. This study aimed to develop a robust human activity recognition (HAR) system to improve continuous physical activity recognition in the context of hip fracture rehabilitation. 24 healthy older adults aged $\geq$ 80 years were included as an age-matched feasibility cohort. Participants performed activities of daily living under simulated free-living conditions for 75 minutes while wearing two accelerometers positioned on the lower back and anterior upper thigh. To address and enhance model robustness to inter-person variability, a novel synthetic data generation scheme was proposed to capture common gait characteristics across individuals. Our real data and synthetic data have been published along with this work. Leveraging this synthetic dataset, a feature selection pipeline was subsequently applied to identify a compact set of six discriminative features for classifying walking, standing, sitting, lying down, and postural transfers. Model robustness was evaluated using leave-one-subject-out cross-validation. The synthetic data demonstrated potential to improve generalization across participants. The resulting feature intervention model (FIM), aided by synthetic data guidance, achieved reliable activity recognition with mean F1-scores of 0.896 $\pm$ 0.100 for walking, 0.927 $\pm$ 0.039 for standing, 0.997 $\pm$ 0.004 for sitting, 0.937 $\pm$ 0.202 for lying down, and 0.816 $\pm$ 0.120 for postural transfers. Compared with a control condition model without synthetic data, the FIM significantly improved the postural transfer detection, i.e., an activity class of high clinical relevance that is often overlooked in existing HAR literature. In conclusion, these preliminary results demonstrate the feasibility of robust activity recognition in older adults. Further validation in hip fracture patient populations is required to assess the clinical utility of the proposed monitoring system. This paper has been submitted to Nordic-Digihealth 2026 conference for review.

\end{abstract}
\section{Introduction}
Hip fractures pose a growing burden on healthcare systems due to population ageing~\cite{kannus1996epidemiology}. Although surgical treatment is often successful, more than half of patients fail to regain prefracture mobility within one year~\cite{bertram2011review,vochteloo2013more}, resulting in reduced independence in activities of daily living (ADL)~\cite{pasco2005human} and long-term declines in health-related quality of life (HRQoL)~\cite{gjertsen2016quality}. Increasing physical activity during rehabilitation is one of the few modifiable strategies shown to improve mobility and ADL independence~\cite{taraldsen2014physical,fitzgerald2018mobility,davenport2015physical,willems2017physical}, and to reduce secondary fracture risk by preventing sarcopenia and balance impairments~\cite{taylor2014physical,daskalopoulou2017physical,fries1996physical,fielding2011sarcopenia}. Despite this clinical importance, physical activity is rarely quantified in routine care~\cite{talkowski2009patient}. Wearable activity trackers have therefore been introduced in postoperative hip fracture rehabilitation~\cite{resnick2011physical,fleig2016sedentary}, yet their accuracy remains uncertain, as they perform poorly for low-intensity movements such as slow or shuffling gait, which are prevalent in geriatric patients~\cite{matthews2002sources,fleig2016sedentary}.

Recent validation work reinforces these concerns. In our previous study~\cite{vandartel2021feasibility}, a commercial human-activity-recognition (HAR) algorithm~\cite{annegarn2011objective} substantially underestimated walking time in hip-fracture patients aged 82 $\pm$ 6 years, frequently misclassifying walking as standing. Such errors arise because most HAR algorithms are calibrated on middle-aged adults, whereas older adults present slower gait and altered movement amplitudes~\cite{bai2022are,zhang2011continuous,iosa2014development}. As a result, HAR models developed in younger populations often fail to capture ambulation behavior in geriatric users~\cite{delrosario2014comparison}.

Accurate monitoring also requires recognizing the functional milestones that determine clinical discharge readiness, including sitting, standing, lying down, walking, and transfers (sit-to-stand, stand-to-sit, sit-to-lie, lie-to-sit)~\cite{lee2020postoperative}. Yet most HAR studies involving older adults overlook transfers~\cite{davoudi2021effect,pannurat2017analysis,chernbumroong2013elderly,sekine2000classification,bai2022are}. The only system addressing all milestones, developed by Allen et al.~\cite{allen2006classification}, was not evaluated in free-living conditions (FLC), where individual habits and environmental unpredictability introduce substantial intra-activity variation~\cite{straczkiewicz2021systematic}.

Collecting data under FLC introduces its own challenges. Because participants perform activities at self-selected frequency and duration, sensor data become highly imbalanced~\cite{bulling2014tutorial}. Imbalances can limit the representation of diverse gait patterns and reduce robustness to inter-person variability—one of the major determinants of generalization failure in HAR~\cite{dehghani2019quantitative,ramasamy2018recent,akila2019highly,ferrari2020personalization}. This indicates the following methodological gap: HAR development for geriatric monitoring must account simultaneously for free-living variability and inter-subject gait differences.

To address these gaps, we developed a new HAR algorithm for detecting hospital-discharge-relevant activities in the context of geriatric hip fracture rehabilitation. Specifically, we developed the HAR system on an age-appropriate cohort of healthy older adults aged $\geq$ 80 years, whose gait characteristics better reflect the target clinical population~\cite{haleem2008mortality}. Our study design captured intra-activity variation by collecting wearable-sensor data under simulated FLC rather than scripted laboratory tasks. To improve robustness to inter-person gait variability and reduce overfitting to idiosyncratic patterns under realistic free-living class imbalance, we proposed a novel synthetic data generation scheme, which captured common movement characteristics across individuals. Rather than explicitly rebalancing activity classes, class-specific synthetic data were generated to enrich the representation of shared gait patterns within each activity, thereby increasing the effective diversity of underrepresented classes and mitigating class-imbalance-induced overfitting in the feature selection process. The synthetic data generation scheme was developed to support more reliable physical activity monitoring during hip fracture rehabilitation in the future.

\section{Methods}
\subsection{Data collection}
This prospective observational study was conducted at the eHealth House (eHH) of the TechMed Simulation Centre, University of Twente (Enschede, The Netherlands). The dataset is available upon request via Zenodo (restricted access, DOI: 10.5281/zenodo.18222937). The eHH is a controlled environment designed to simulate free-living conditions (FLC) and includes a living room, kitchen, bedroom, and bathroom. All experimental sessions were recorded using five fixed cameras installed throughout the eHH. Participants were eligible if they were aged 80 years or older and physically able to participate independently. Individuals with cognitive impairments or mobility disorders that impeded task performance were excluded. According to Dutch law, and based on a ruling from the Medical Research Ethics Committee (MREC) Arnhem–Nijmegen, the study was exempt from the Medical Research Involving Human Subjects Act. Ethical approval was granted by the Ethics Committee of the Natural Sciences and Engineering Sciences of the University of Twente. All participants provided written informed consent prior to participation.

Each participant visited the eHH once for a 75-minute session. During the visit, participants performed a set of activities of daily living (ADL) at their own pace and in their preferred order. These tasks reflected functional discharge criteria for geriatric hip fracture patients and included: (1) walking inside the eHH (between the living room, kitchen, bedroom, and front door), (2) visiting the toilet once, (3) getting in and out of bed once (sit-to-lie, lying, and lie-to-sit), (4) preparing a meal (walking to the kitchen, cutting food, and returning to the living room), and (5) preparing a drink (walking to the kitchen, pouring a drink, and returning to the living room).

\begin{figure}
    \centering
    \includegraphics[width=0.8\linewidth]{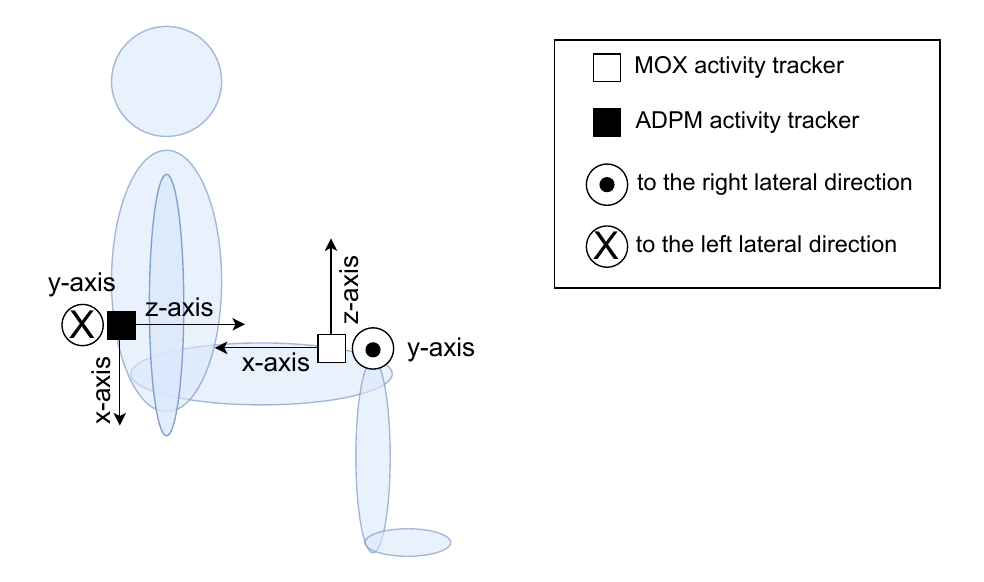}
    \caption{Schematic of activity trackers’ placements on the upper thigh and lower back.}
    \label{fig:sensorplacement}
\end{figure}

Participants’ movements and postures were recorded using two wearable devices: the MOX activity monitor (Maastricht Instruments, The Netherlands) and the APDM activity tracker (Hankamp Rehab BV, The Netherlands). The MOX is a waterproof device containing a single triaxial accelerometer and was attached to the upper thigh, approximately 10 cm above the knee, using a medical plaster. Data were recorded at 25 Hz. This placement was selected based on evidence that upper-leg accelerations exhibit low inter-person variability, facilitating better generalization in HAR~\cite{janidarmian2017comprehensive}. The APDM comprises a triaxial accelerometer, gyroscope, and magnetometer, and was worn on the lower back using a strap. Data were recorded at 128 Hz. This location provides robust measurements of sedentary behaviors with low sensitivity to postural differences~\cite{yang2010review}, and captures representative whole-body motion due to its proximity to the body’s center of mass. The combination of thigh and lower-back sensors (Fig.~\ref{fig:sensorplacement}) was considered necessary to reliably distinguish all static and dynamic activities, as a single sensor location was insufficient.

To obtain gold-standard labels, two independent reviewers annotated the video recordings into the following activity classes: walking, standing, sitting, lying (supine, left lateral recumbent, right lateral recumbent), and transfers (stand-to-sit, sit-to-stand, sit-to-lie, and lie-to-sit). Disagreements were resolved by a third reviewer. The activity annotations were synchronized with the sensor recordings by aligning the timestamps of the first annotated sit-to-stand transfer observed in the videos with the corresponding acceleration signals.

After synchronization, all subsequent analyses were performed using accelerometer data only. The gyroscope and magnetometer signals were excluded, as previous studies showed that accelerometers alone outperform multimodal sensor combinations for comparable HAR tasks, with no significant performance gains from sensor fusion~\cite{shoaib2014fusion,shoaib2016complex}. All data processing was conducted in MATLAB (R2022a, MathWorks Inc., Natick, MA, USA).

\subsection{Models}\label{sec:models}
To improve robustness to inter-person variability and reduce overfitting to idiosyncratic patterns due to class imbalance, we constructed a class-specific synthetic dataset (see Section~\ref{sec:syntheticdata}) representing common gait patterns across individuals. Using this dataset, an HAR model incorporating synthetic data was proposed and compared against a control condition model trained solely on real data.

\begin{figure}
    \centering
    \includegraphics[width=0.6\linewidth]{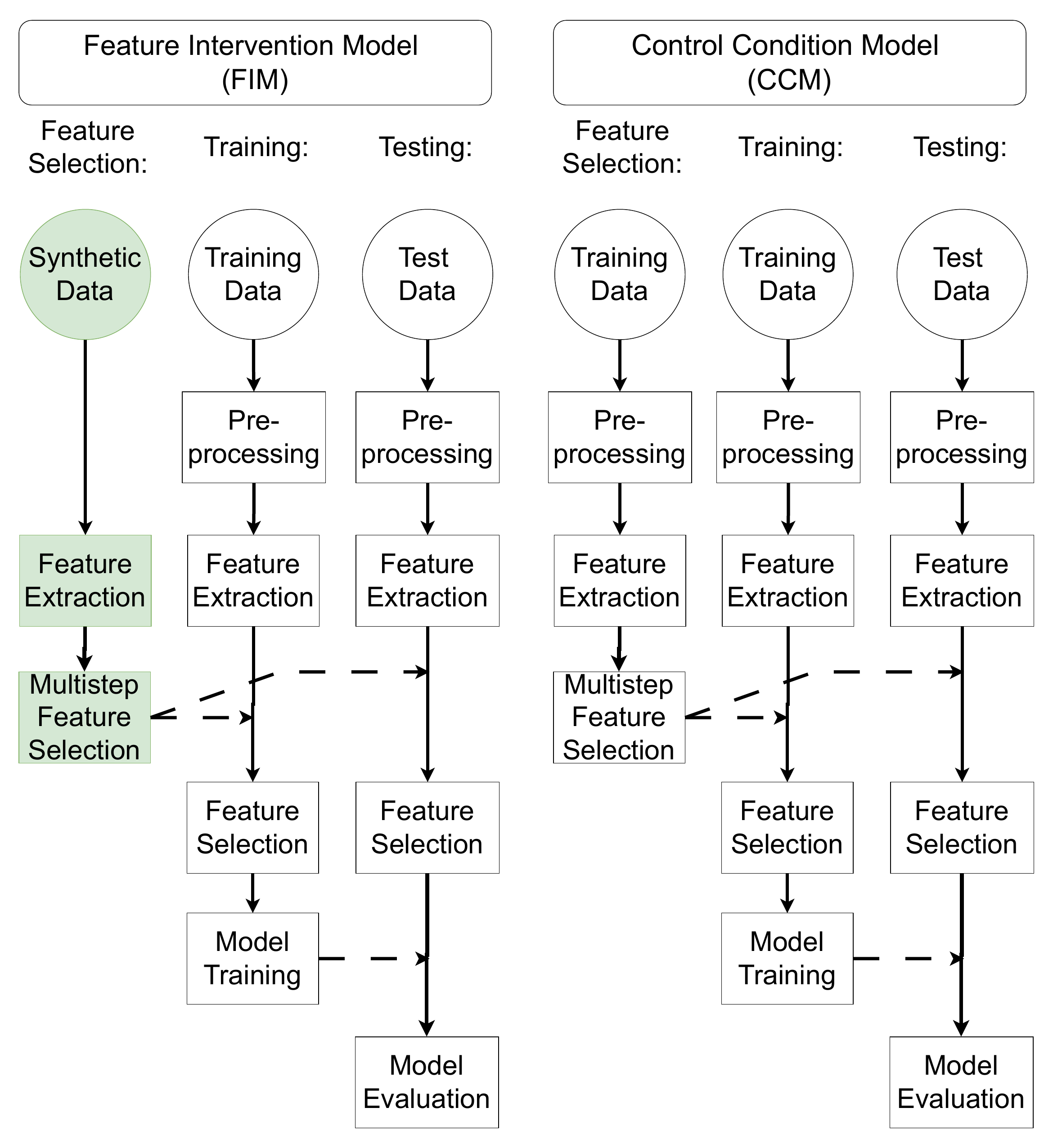}
    \caption{General workflow of three proposed human activity recognition models. The process blocks involving the direct usage of synthetic data or derived features are colored in green.}
    \label{fig:models}
\end{figure}

The synthetic data-guided model, termed the feature intervention model (FIM) (left flow chart in Fig.~\ref{fig:models}), examined whether synthetic data could better guide the identification of generalizable features than the real data. Specifically, for the FIM model, feature selection was performed solely on the complete synthetic dataset while excluding the real data. This design choice was motivated by the hypothesis that the synthetic data suppresses individual-specific variability, thereby encouraging the feature selection process to capture fundamental, subject-invariant activity characteristics rather than incidental or participant-specific patterns, even in the presence of class imbalance. Combining features selected from both real and synthetic data was intentionally avoided, since features identified from real data may encode subject-specific or dataset-specific biases, which could dilute the intended regularization effect of the synthetic data and compromise robustness to inter-person variability. After feature selection on synthetic data, the selected features were extracted from the real data. Model training and evaluation were subsequently conducted on the real data using leave-one-subject-out cross-validation. The control condition model (CCM) (right flow chart in Fig.~\ref{fig:models}) followed the standard activity recognition chain (ARC) using only real data. Feature selection was performed on 25\% of the real data, after which the selected features were used to train and evaluate the model on the remaining 75\% using leave-one-subject-out cross-validation.

Both HAR models followed the standard activity recognition chain (ARC)~\cite{bulling2014tutorial}, comprising preprocessing, feature extraction, feature selection, model training, and model evaluation. Each component is described in detail in the following sections.

\subsection{Preprocessing}\label{sec:preprocessing}
Random high-frequency noise was suppressed by smoothing all physical-activity signals using a Savitzky–Golay filter~\cite{savitzky1964smoothing}. This filter was preferred over a moving average due to its superior ability to preserve peak shape and amplitude during smoothing~\cite{schafer2011frequency}. The filter was configured with a frame length of 0.12~s and a second-order polynomial. The selected frame length was previously shown to be suitable for smoothing gait signals in older adults~\cite{bijnens2019optimization}, while a polynomial order of 2 was sufficient to capture local movement characteristics in accelerometer data~\cite{lee2018enhanced}.

After filtering, signals were segmented into non-overlapping 2-second windows. Non-overlapping windows were chosen because overlapping windows increase computational cost without improving generalization performance in HAR systems~\cite{dehghani2019quantitative}. A window length of 2 seconds was selected as it has been shown to yield low classification error rates across a wide range of physical activities~\cite{banos2014window}, and is particularly suitable for older adults~\cite{bijnens2019optimization}. Windows containing mixed activity annotations were excluded to avoid classification ambiguity~\cite{atallah2011sensor}.

\subsection{Synthetic data generation}\label{sec:syntheticdata}
Synthetic data were generated by extracting generalized gait representations that capture common characteristics across individuals while minimizing participant-specific biases. The synthetic dataset is available upon request via Zenodo (restricted access, DOI: 10.5281/zenodo.18230770). First, raw acceleration signals were visually inspected for anomalies. For walking, large baseline drifts along the APDM z-axis were observed in a small number of participants. As no kinematically plausible explanation could be identified, these drifts were removed using a noncausal fourth-order high-pass filter with a cut-off frequency of 0.5~Hz.

Second, generalized activity representations were generated using dynamic time warping barycentre averaging (DBA)~\cite{petitjean2011global}. DBA is a DTW-based averaging technique that aligns time series with varying temporal dynamics by nonlinear stretching or compression, while enforcing alignment of the first and last samples~\cite{muller2007information}. After alignment, the barycentre of the aligned signals is computed, yielding a representative time series that captures common temporal patterns despite inter-individual variability~\cite{petitjean2011global,petitjean2012summarizing,hachaj2018averaging,chan2018evaluation,zhao2020time}. Synthetic windows were generated for walking, standing, sitting, supine lying, left and right lateral recumbent lying, sit-to-stand, stand-to-sit, sit-to-lie, and lie-to-sit transfers. To ensure a fair comparison, the number of 2-second windows per activity was matched between the real and synthetic datasets, preserving identical class imbalance conditions.  

\begin{figure}
    \centering
    \includegraphics[width=\linewidth]{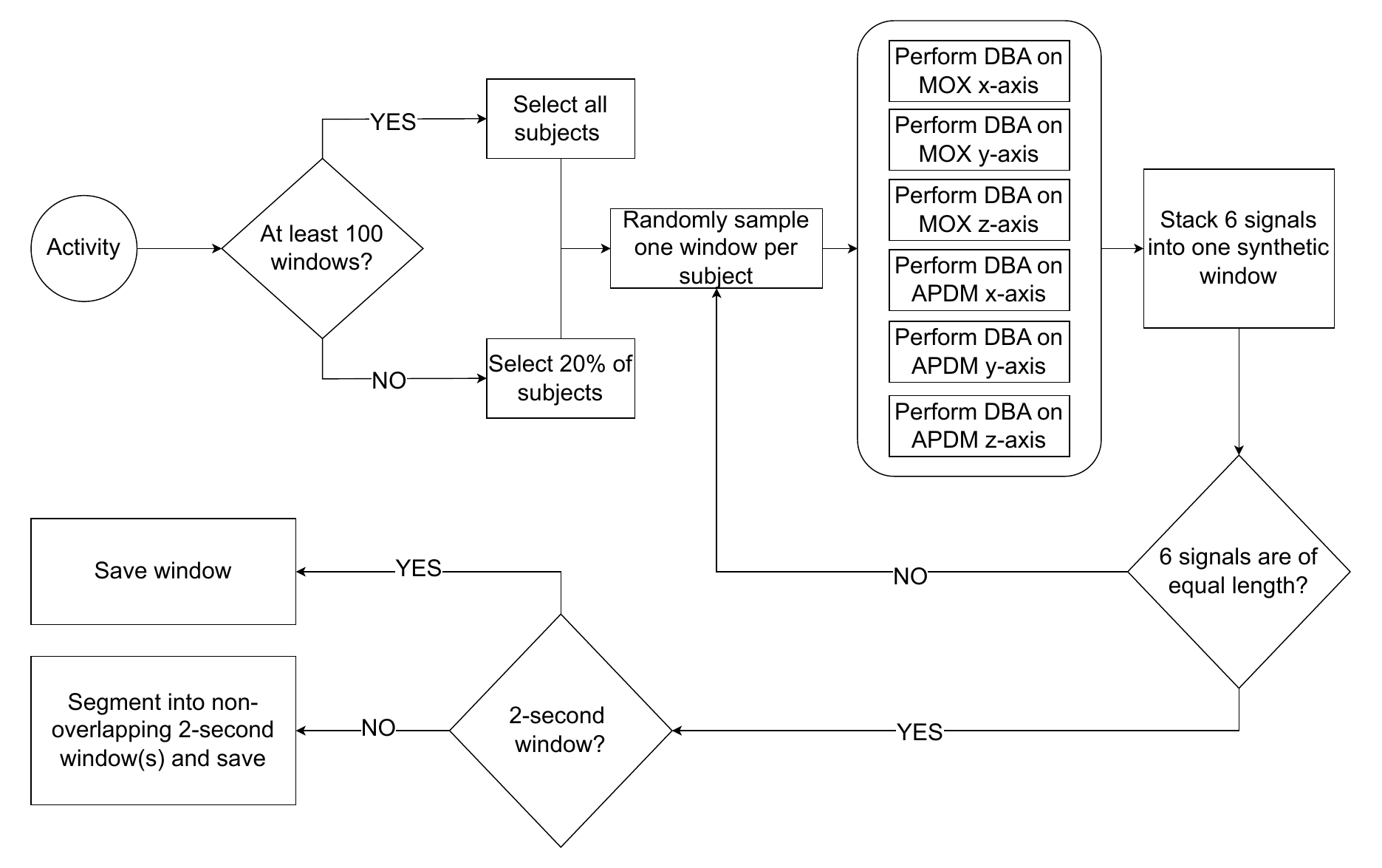}
    \caption{General protocol for the generation of synthetic data using dynamic time warping barycentre averaging (DBA), on both MOX and APDM sensory data.}
    \label{fig:DBA_pipeline}
\end{figure}

A summary of the custom DBA protocol is shown in Fig.~\ref{fig:DBA_pipeline}. The protocol consisted of four steps. First, for each activity, data availability was assessed. If at least 100 windows were available, one window per participant (i.e., 24 windows from 24 participants) was randomly sampled per DBA iteration. Otherwise, a subset-based sampling strategy was adopted: windows were randomly sampled from 20\% of the participants (6 windows per iteration), a proportion chosen empirically, to increase combinatorial diversity, as 134,596 unique combinations exist when selecting 6 participants from 24.

Second, one window was randomly sampled per selected participant. Each window comprised synchronized MOX and APDM signals along the x-, y-, and z-axes. Window length was activity-dependent to reduce DTW boundary artifacts. For walking, 4-second windows were used to exploit periodicity, after which the central 2 seconds were retained. Transfers were processed from start to end due to their variable duration, while static activities were processed using fixed 2-second windows without concern for boundary effects.

Third, DBA was applied separately to each axis of both sensors, resulting in six signals per synthetic window. For transfers, temporal consistency across the six signals was verified, as DTW can yield outputs of unequal length. If inconsistencies were detected, the sampling process was repeated.

Finally, post-processing was applied to obtain 2-second windows. For walking, the first and last seconds of the 4-second window were discarded. Transfer signals longer than 2 seconds were segmented into non-overlapping 2-second windows. No post-processing was required for static activities, as their windows were already of fixed length.

\subsection{Feature extraction}
Handcrafted features were used instead of deep learning–based automated feature extraction for two reasons. First, deep learning methods require large datasets to achieve robust generalization~\cite{logacjov2021harth}, which was not feasible in this study. Second, a primary target of this work was the detection of transfer activities, which occur infrequently in HAR datasets and are therefore poorly suited to data-hungry deep learning approaches~\cite{reyes2016transition,uddin2016random,chen2020method}.

The candidate feature set was selected based on prior HAR literature. Although both time- and frequency-domain features are commonly employed~\cite{figo2010preprocessing}, only time-domain features were considered in this study. Frequency-domain features incur higher computational cost without consistent evidence of superior discriminative performance~\cite{janidarmian2017comprehensive,allen2006classification,elsts2018board}. Moreover, they are primarily effective for quasi-periodic movements with distinct frequency content~\cite{allen2006classification}. Since most activities in this study were static or aperiodic, frequency-domain features were expected to contribute limited additional information and were therefore excluded.

To further reduce multicollinearity, redundant time-domain features were preemptively removed based on domain knowledge. For example, pairs of features such as mean and median, or variance, standard deviation, and mean absolute deviation, were expected to encode largely overlapping information. This pruning resulted in a total of 62 candidate features (Table~\ref{tab:feature_candidates}). Such pre-selection of features is recommended to improve the reliability of subsequent feature selection algorithms~\cite{guyon2003introduction,wallisch2021selection} and to reduce instability in feature optimization procedures~\cite{sanchez2007filter}.

\begin{table}[htbp]
\footnotesize
\centering
\caption{Overview of the time domain features before selection. Each feature was computed for both the MOX and APDM signals across a window of size $N$. Except for the axial correlations and signal vector magnitude, all features were computed along each acceleration axis $a\in\{x,y,z\}$. This resulted in a total of 62 features.}
\label{tab:feature_candidates}

\renewcommand\theadfont{\normalsize\bfseries}
\renewcommand\cellalign{tl}

% Three columns: Feature (X), Formula (X), References (p{0.18\textwidth})
\begin{tabularx}{\textwidth}{>{\raggedright\arraybackslash}X >{\raggedright\arraybackslash}X p{0.18\textwidth}}
\toprule
\textbf{Feature and interpretation} & \textbf{Formula} & \textbf{References} \\
\midrule

The \textbf{mean} provides postural information.
&
\makecell[tl]{$\mu_a = \dfrac{1}{N}\sum_{i=1}^N a_i$}
&
\cite{janidarmian2017comprehensive,bijnens2019optimization,banos2014window,bayat2014study,cleland2013optimal,erdacs2016integrating,lau2010movement,mannini2013activity,olguin2006human,pannurat2017analysis,trost2014machine,chen2021improving}
\\[0.6em]

\hline
\addlinespace[0.5em]

The \textbf{standard deviation} provides information about the intensity of a physical activity.
&
\makecell[tl]{$\sigma_a = \sqrt{\dfrac{\sum_{i=1}^N (a_i-\mu_a)^2}{N-1}}$}
&
\cite{janidarmian2017comprehensive,bayat2014study,cleland2013optimal,erdacs2016integrating,lau2010movement,trost2014machine,chen2021improving,pirttikangas2006feature,preece2009activity,baek2004accelerometer}
\\[0.6em]

\hline
\addlinespace[0.5em]

The \textbf{root mean square} is identical to the standard deviation for signals with a mean of zero.
&
\makecell[tl]{$\mathrm{RMS}(a)=\sqrt{\dfrac{1}{N}\sum_{i=1}^N a_i^{\,2}}$}
&
\cite{janidarmian2017comprehensive,bayat2014study,chernbumroong2013elderly,erdacs2016integrating,chen2021improving}
\\[0.6em]

\hline
\addlinespace[0.5em]

\textbf{Axial correlations} distinguish unidirectional vs.\ multidirectional translations (e.g., walking vs.\ stair climbing).
&
\makecell[tl]{$\dfrac{\mathrm{cov}(x,y)}{\sigma_x\sigma_y}$,\quad
$\dfrac{\mathrm{cov}(x,z)}{\sigma_x\sigma_z}$,\quad
$\dfrac{\mathrm{cov}(y,z)}{\sigma_y\sigma_z}$}
&
\cite{arif2015physical,bao2004activity,bayat2014study,chernbumroong2013elderly,cleland2013optimal,ravi2005activity,trost2014machine,janidarmian2017comprehensive,chen2021improving}
\\[0.6em]

\hline

The \textbf{minimum} describes the lowest value measured along a single sensing axis.
&
\makecell[tl]{$\min(a_1, a_2, \dots, a_N)$}
&
\cite{chernbumroong2013elderly,erdacs2016integrating,mannini2013activity,pannurat2017analysis,janidarmian2017comprehensive}
\\[0.6em]

\hline

The \textbf{maximum} describes the highest value measured along a single sensing axis.
&
\makecell[tl]{$\max(a_1, a_2, \dots, a_N)$}
&
\cite{chernbumroong2013elderly,erdacs2016integrating,mannini2013activity,pannurat2017analysis,janidarmian2017comprehensive}
\\[0.6em]

\hline

The \textbf{interquartile range} separates sedentary behaviours from physical activities.
&
$Q_3 - Q_1$
&
\cite{janidarmian2017comprehensive,chen2021improving}
\\[0.6em]

\hline
\addlinespace[0.5em]

The \textbf{skewness} describes the distribution asymmetry of movement accelerations.
&
\makecell[tl]{$\dfrac{\sum_{i=1}^N (a_i-\mu_a)^3}{(N-1)\,\sigma_a^3}$}
&
\cite{cleland2013optimal,erdacs2016integrating,baek2004accelerometer,trost2014machine,janidarmian2017comprehensive,chen2021improving}
\\[0.6em]

\hline
\addlinespace[0.5em]

The \textbf{kurtosis} describes the likelihood of extreme values in a movement signal.
&
\makecell[tl]{$\dfrac{\sum_{i=1}^N (a_i-\mu_a)^4}{(N-1)\,\sigma_a^4}$}
&
\cite{cleland2013optimal,erdacs2016integrating,baek2004accelerometer,trost2014machine,janidarmian2017comprehensive,chen2021improving}
\\[0.6em]

\hline
\addlinespace[0.5em]

The \textbf{mean-crossing rate} captures how often accelerations change sign relative to the mean.
&
\makecell[tl]{%
$\begin{array}{l}
\dfrac{1}{2}\sum_{i=2}^N \Big|\, \operatorname{sgn}(a_i-\mu_a) \\
- \operatorname{sgn}(a_{i-1}-\mu_a) \Big|
\end{array}$
}
&
\cite{arif2015physical,banos2014window,trost2014machine,chen2021improving}
\\[0.6em]

\hline
\addlinespace[0.5em]

The \textbf{signal vector magnitude} distinguishes sedentary from active behaviours. All accelerations were high-pass filtered (noncausal, 4th-order, 0.5\,Hz cutoff) beforehand.
&
\makecell[tl]{$\dfrac{1}{N}\sum_{i=1}^N \sqrt{x_i^{\,2}+y_i^{\,2}+z_i^{\,2}}$}
&
\cite{janidarmian2017comprehensive,figo2010preprocessing}
\\
\bottomrule
\end{tabularx}
\end{table}

\subsection{Multistep feature selection}
\begin{figure}
    \centering
    \includegraphics[width=\linewidth]{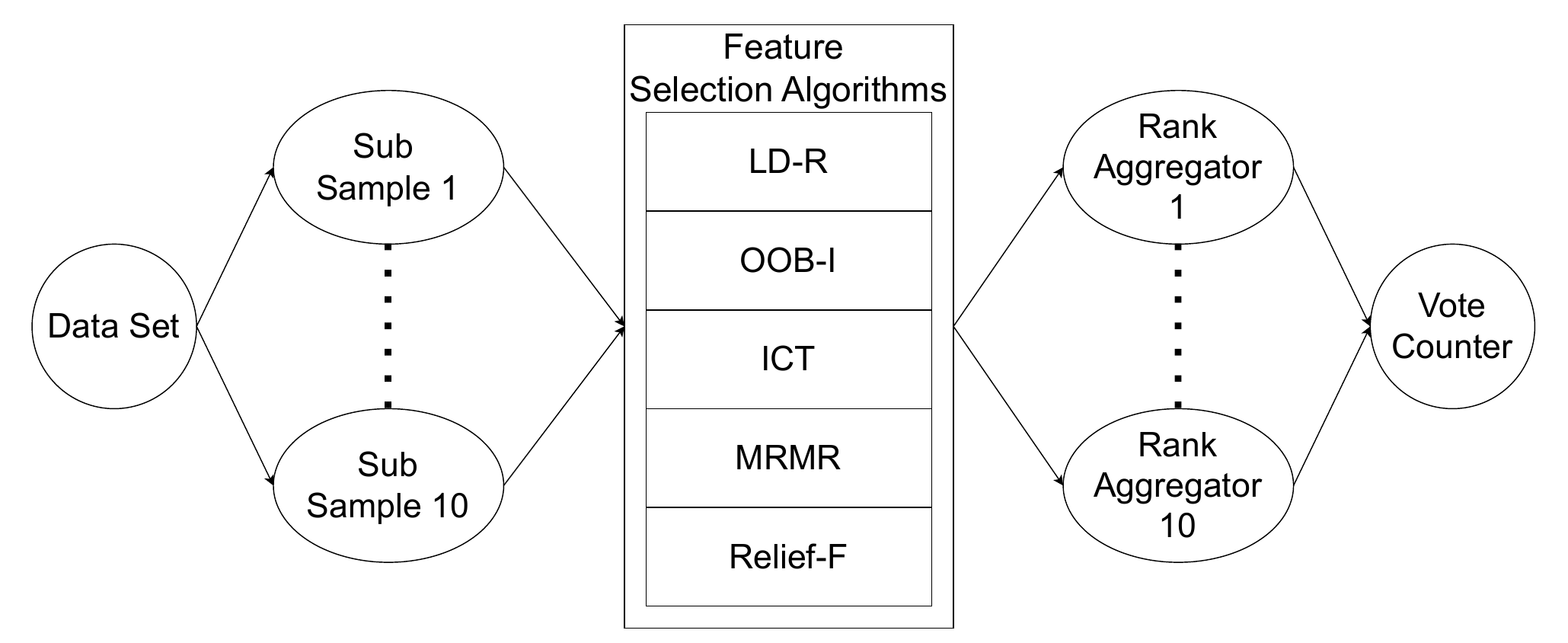}
    \caption{Overview of the multistep feature selection (FS) pipeline. The pipeline was applied to the real data to build the control condition model, and it was applied to the synthetic data to build the complete data intervention model and the feature intervention model. Five FS algorithms were selected in this work, including Relief-F~\cite{urbanowicz2018relief}, maximum relevance minimum redundancy (MRMR)~\cite{ding2011minimum}, interaction-curvature tests embedded into a decision tree (ICT)~\cite{loh2002regression}, out-of-bag feature importance by permutation embedded into a random forest (OOB-I)~\cite{breiman2001random}, and regularization embedded into a linear discriminant (LD-R)~\cite{guo2007regularized}.}
    \label{fig:FSpipeline}
\end{figure}

To improve the generalizability of the HAR models, a feature selection (FS) pipeline was developed to identify a minimal and robust subset of features from the initial candidate set. The pipeline, applied identically to real and synthetic data, is summarized in Fig.~\ref{fig:FSpipeline}. It addresses two key challenges in FS.

First, no single FS algorithm is universally optimal, as performance strongly depends on the application and data characteristics~\cite{boloncanedo2019ensembles,seijo-pardo2017testing}. To mitigate this limitation, a heterogeneous feature selection ensemble (HFSE) was employed~\cite{boloncanedo2019ensembles}. By combining multiple FS algorithms with complementary selection heuristics, HFSEs have been shown to produce more generalizable feature sets than individual algorithms~\cite{seijo-pardo2017testing,drotar2019ensemble,bolon-canedo2012ensemble,boloncanedo2019ensembles,benbrahim2018ensemble}. Second, FS results are often sensitive to small variations in the data~\cite{wallisch2021selection}. To improve robustness, FS stability was imposed as a secondary criterion. Following~\cite{kalousis2007stability}, the FS procedure was repeated 10 times on activity-stratified subsamples with approximately 78\% overlap. The resulting feature sets were aggregated by vote counting: a feature was retained if it was selected in at least 5 out of 10 repetitions. This threshold is commonly used to eliminate features selected by chance~\cite{bunea2011penalized,blackstone2001breaking,salarbaks2020pneumonia}.

The HFSE design consisted of two steps: (1) selecting the constituent FS algorithms, and (2) aggregating their ranking outputs.

For the first step, five FS algorithms were included: Relief-F~\cite{urbanowicz2018relief}, maximum relevance minimum redundancy (MRMR)~\cite{ding2011minimum}, interaction–curvature tests embedded in a decision tree~\cite{loh2002regression}, out-of-bag permutation importance from a random forest (OOB-I)~\cite{breiman2001random}, and regularized linear discriminant analysis (LD-R)~\cite{guo2007regularized}. Algorithm selection was guided by two criteria: diversity of selection heuristics and algorithmic stability~\cite{bolon-canedo2012ensemble,boloncanedo2019ensembles}. Diversity ensured that the strengths of one algorithm compensated for the limitations of another (Table~\ref{tab:selection_algorithms}). For example, Relief-F captures feature interactions but does not address redundancy, whereas MRMR explicitly penalizes redundant features but ignores interactions.

\begin{table}[htbp]
\footnotesize
\centering
\caption{Overview of the feature selection algorithms used in the heterogeneous feature selection ensemble, with their properties highlighted. NA indicates not applicable.}
\label{tab:selection_algorithms}
\resizebox{\textwidth}{!}{
\begin{tabular}{l l l l l}
\toprule
\textbf{Algorithm} & \textbf{Type} & \textbf{Relationship} & \textbf{Feature redundancy} & \textbf{Classifier type} \\
\midrule
Relief-F & Filter   & Multivariate & Not accounted for & NA \\
MRMR     & Filter   & Univariate   & Accounted for     & NA \\
ICT\tnote{a} & Embedded & Multivariate & NA & Nonlinear \& greedy \\
OOB importance\tnote{b} & Embedded & Univariate & NA & Nonlinear \& non-greedy \\
Regularisation\tnote{c} & Embedded & Univariate & NA & Linear \\
\bottomrule
\end{tabular}
}
\begin{tablenotes}
\footnotesize
\item MRMR: maximum relevance minimum redundancy, ICT: interaction-curvature tests, OOB: out-of-bag.
\item[a] Embedded into a decision tree classifier,
\item[b] Embedded into a random forest classifier,
\item[c] Embedded into a linear discriminant classifier.
\end{tablenotes}
\end{table}

Feature selection should identify features that reflect true activity characteristics rather than artifacts of a particular data split. We therefore assessed algorithm stability by examining how consistently each feature selection method selected the same features across repeated subsamples of the data. Algorithm stability was quantified using the Tanimoto similarity $T$. Each FS algorithm was applied to 10 activity-stratified subsamples of the synthetic data, producing 10 ranking lists. Following standard practice~\cite{kalousis2007stability,boloncanedo2019ensembles,chandrashekar2014survey}, the top 10 features from each ranking were used to form feature subsets, and $T$ was computed for all 45 unique subset pairs (i.e., the number of unique unordered pairs that can be formed from 10 subjects):

\begin{equation}
    \label{eq:tanimoto}
    T(s,s')=1-\dfrac{|s|+|s'|-2|s\cap s'|}{|s|+|s'|-|s\cap s'|}
\end{equation}

where $s$ and $s'$ denote two feature subsets, $|\cdot|$ their cardinality, and $s\cap s'$ their intersection. $T=0$ indicates no overlap and $T=1$ identical subsets. The stability of an FS algorithm was defined as the average $T$ across all subset pairs.

For the second step, robust rank aggregation (RRA)~\cite{kolde2012robust} was used to combine the feature rankings. Each feature received five ranks, i.e., one from each FS algorithm, ranging from 1 to $m$, where $m$ is the total number of features. Ranks were normalized to $r_i\in(0,1]$, yielding $\bm{r}=(r_1,\dots,r_5)$, which was reordered such that $r_{(1)}\leq\dots\leq r_{(5)}$.

RRA evaluates the probability of observing a rank that is $r_{(k)}$ or smaller under the null hypothesis that all ranks are independently drawn from a uniform distribution $\mathcal{U}(0,1)$. This probability is given by the binomial distribution

\begin{equation}
    \label{eq:binomial_order}
    \beta_{(k)}:=\sum_{\ell=k}^5\binom{5}{\ell}r_{(k)}^\ell\left(1-r_{(k)}\right)^{(5-\ell)}
\end{equation}

\begin{equation}
    \label{eq:exact_p_RRA}
    p(\bm{r}) = \min\left(1, 5\cdot \underset{k\in\{1,2,3,4,5\}}{\arg\min} \beta_{(k)}(\bm{r})\right)
\end{equation}

which includes a Bonferroni correction. Features with p-values $<$ 0.05 were identified as consistently informative across FS algorithms.

\subsection{Model training}
For the CCM and FIM, the same classification algorithm and hyperparameters were used to ensure fair comparisons. The k-nearest neighbors (KNN) classifier was selected because it has consistently demonstrated strong performance across HAR studies~\cite{janidarmian2017comprehensive,lau2010movement,pirttikangas2006feature,gupta2014feature}, typically requires fewer features and shorter window lengths to achieve high accuracy~\cite{banos2014window}, and remains robust across different sensor placements~\cite{janidarmian2017comprehensive}.

Hyperparameter tuning was not performed due to severe class imbalance and limited data availability for minority classes, which made partitioning the data into reliable tuning sets infeasible. Instead, hyperparameters were selected based on established findings in the literature. Prior studies indicate that values of $k$ in the range [3, 10] yield comparable performance across diverse HAR tasks, with minimal sensitivity to the exact choice~\cite{janidarmian2017comprehensive,shoaib2016complex,banos2014window,atallah2011sensor,gupta2014feature,bennasar2022significant}. Based on these findings, $k=5$ was selected. Smaller values of $k$ were avoided due to increased overfitting risk~\cite{james2013introduction}, while excessively large values reduce KNN’s sensitivity to local structure~\cite{janidarmian2017comprehensive}. An odd value was preferred to prevent ties in majority voting~\cite{shoaib2016complex}. The Euclidean distance metric was used, as it has been shown to perform optimally in extensive KNN evaluations for HAR~\cite{janidarmian2017comprehensive}. Because KNN is sensitive to feature scale, all features were Z-standardized prior to classification to ensure equal contribution across feature dimensions~\cite{gupta2014feature}.

All models were trained to recognize five activity classes: walking, standing, sitting, lying, and transfers. Supine, left, and right lateral recumbent positions were merged into a single lying class, as they are functionally equivalent and difficult to distinguish reliably under free-living conditions due to sensor orientation variability. Due to limited samples for individual transfer types (sit-to-stand, stand-to-sit, sit-to-lie, and lie-to-sit), these were also merged into a single transfer class.

\subsection{Evaluation}
Model performance was quantified using the F1-score, defined as:

\begin{equation}
    \label{eq:f1_score}
    \begin{split}
        & \text{F1-score} = 2\cdot\dfrac{\text{precision}\cdot\text{recall}}{\text{precision + recall}} \\
        & \text{precision} = \dfrac{\text{true positives}}{\text{true positives + false positives}} \\
        & \text{recall} = \dfrac{\text{true positives}}{\text{true positives + false negatives}}
    \end{split}
\end{equation}

Differences in participant-level F1-scores across models were evaluated using the Friedman test with Wilcoxon signed-rank post-hoc comparisons (Bonferroni-corrected).

\section{Results}
\subsection{Dataset}
An overview of the dataset is provided in Table~\ref{tab:act_freq}. A total of 24 participants were included in the study (11 males, 13 females). The cohort had a median age of 82 years (interquartile range: 81–85 years). Data collection under simulated free-living conditions resulted in pronounced imbalance at multiple levels. First, substantial class imbalance was observed, with transfers constituting the minority class (1.4\%) and sitting the majority class (73.5\%). Second, participation imbalance was present across activities; for example, only five and two participants contributed data for left and right lateral recumbent positions, respectively. Third, the number of observations per participant varied considerably within activities. For right lateral recumbent lying, one participant contributed only 6 windows, whereas another contributed 305 windows.

\begin{table}[H]
\centering
\caption{Overview of the number of 2-second windows for all physical activities.
\# denotes the number of windows per activity, \% the proportion of windows relative to the total dataset, and N the number of participants contributing to each activity. 23*: the sit-to-lie transfer of one participant was missing as the activity could not be verified with the video recordings due to a faulty camera orientation.}
\begin{tabular}{llllll}
\hline
Physical activity       & \multicolumn{2}{l}{Available windows} & N & \multicolumn{2}{l}{Windows per participant} \\
                        & \#                 & \%               &                    & Mean              & Min - Max               \\ \hline
Walking                 & 2,244              & 8.0              & 24                 & 90.5              & 58 - 139                \\ 
Standing                & 3,258              & 11.6             & 24                 & 133               & 36 - 234                \\ 
Sitting                 & 20,677             & 73.5             & 24                 & 967               & 322 - 1,487             \\ 
Supine                  & 929                & 3.3              & 21                 & 19                & 5 - 171                 \\ 
Left lateral recumbent  & 300                & 1.1              & 5                  & 24                & 15 - 206                \\ 
Right lateral recumbent & 311                & 1.1              & 2                  & 155.5             & 6 - 305                 \\ 
Sit-to-stand transfer   & 178                & 0.6              & 24                 & 8                 & 2 - 10                  \\ 
Stand-to-sit transfer   & 163                & 0.6              & 24                 & 7                 & 3 - 11                  \\ 
Sit-to-lie transfer     & 35                 & 0.1              & 23*                & 1                 & 1 - 3                   \\ 
Lie-to-sit transfer     & 41                 & 0.1              & 24                 & 1.5               & 1 - 4                   \\ \hline
\end{tabular}\label{tab:act_freq}
\end{table}

\subsection{Feature selection}
\begin{figure}
    \centering
    \includegraphics[width=\linewidth]{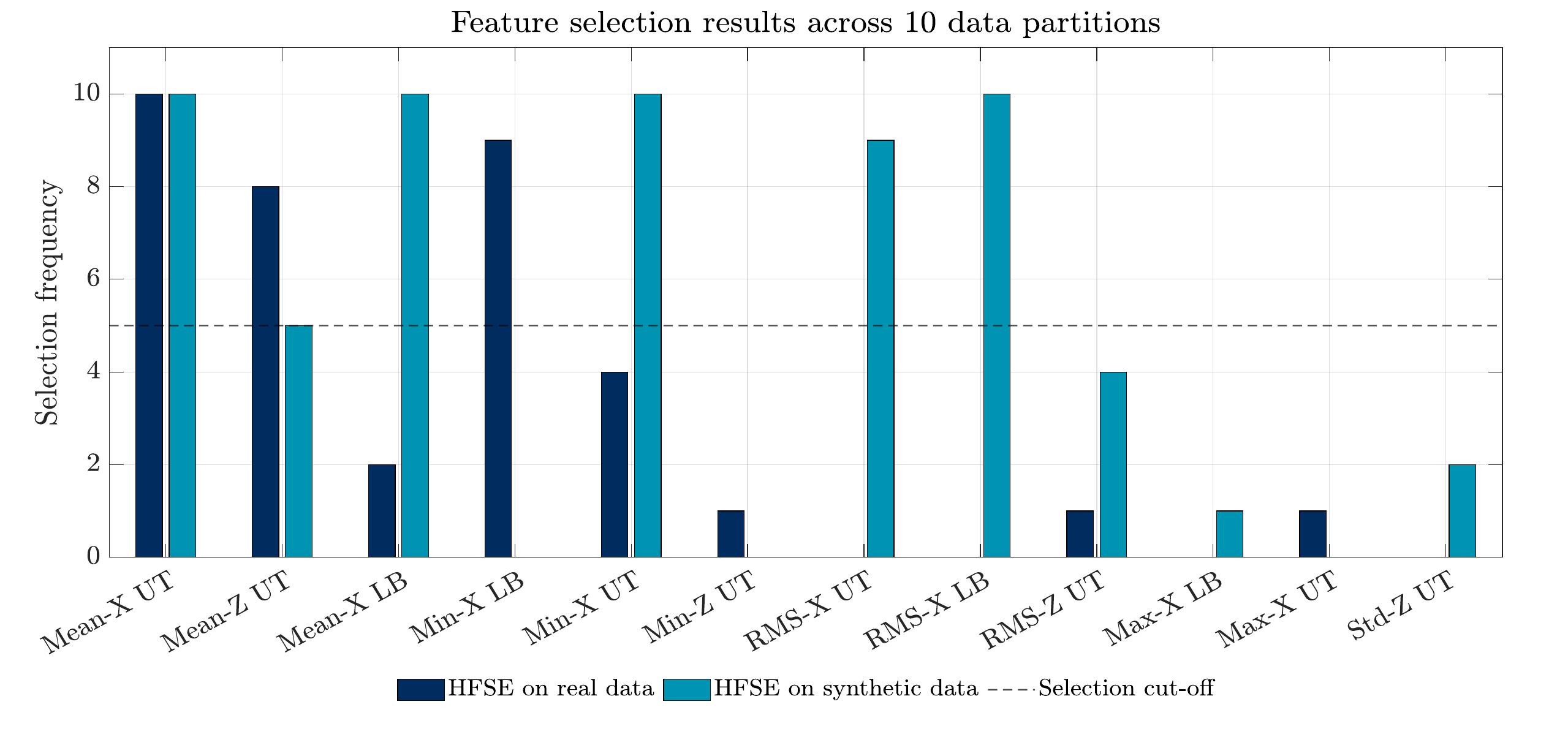}
    \caption{Overview of feature selection results using the heterogeneous feature selection ensemble (HFSE) across the 10 subsamples of the real and synthetic data. X and Z denote the acceleration axes. UT, LB, and RMS are abbreviations for upper thigh, lower back, and root-mean-square, respectively.}
    \label{fig:selection_stability}
\end{figure}

All five individual FS algorithms demonstrated sufficient stability to contribute to the HFSE, with Tanimoto similarity scores ranging from 0.648 to 1.0, indicating good to excellent stability~\cite{boloncanedo2019ensembles}. The HFSE outcomes across 10 subsamples of the real and synthetic datasets are shown in Fig.~\ref{fig:selection_stability}.

For the real data, three features were stably selected across all subsamples: mean accelerations from the upper thigh (Mean-X UT, Mean-Z UT) and the minimum acceleration from the lower back (Min-X LB). For the synthetic data, six features were stably selected: mean accelerations (Mean-X UT, Mean-Z UT, Mean-X LB), minimum acceleration (Min-X UT), and root mean square accelerations (RMS-X UT, RMS-X LB). Among these, only the two upper-thigh mean acceleration features overlapped with those selected from the real data.

\begin{figure}[H]
    \centering
    \includegraphics[width=\linewidth]{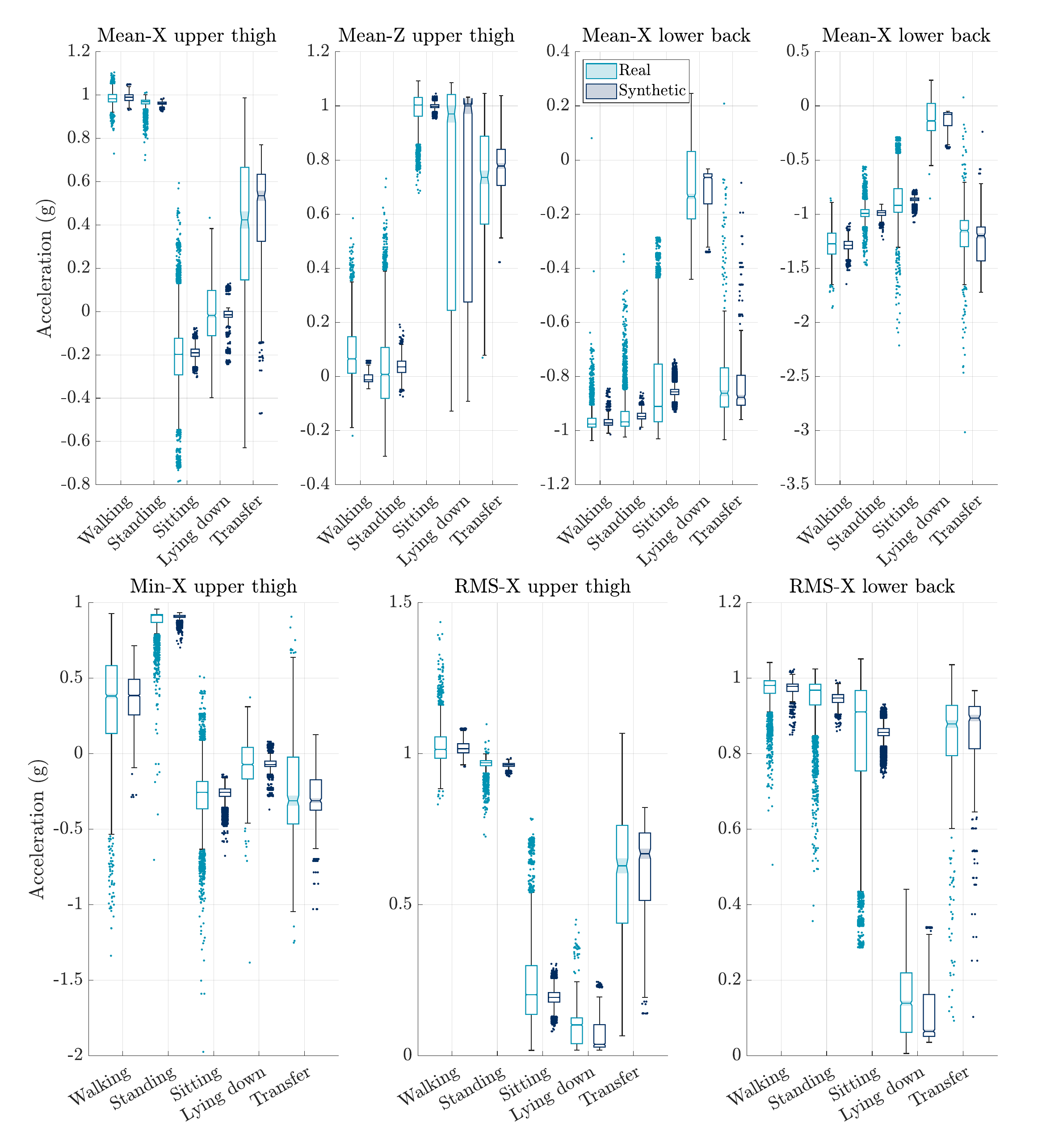}
    \caption{Class-conditional value distributions of the features selected in accordance with the heterogeneous feature selection ensemble pipeline across the complete real and synthetic datasets. X and Z denote the acceleration axes. RMS is the abbreviation for root-mean-square.}
    \label{fig:feature_boxplots}
\end{figure}

\begin{figure}[H]
    \centering
    \includegraphics[width=\linewidth]{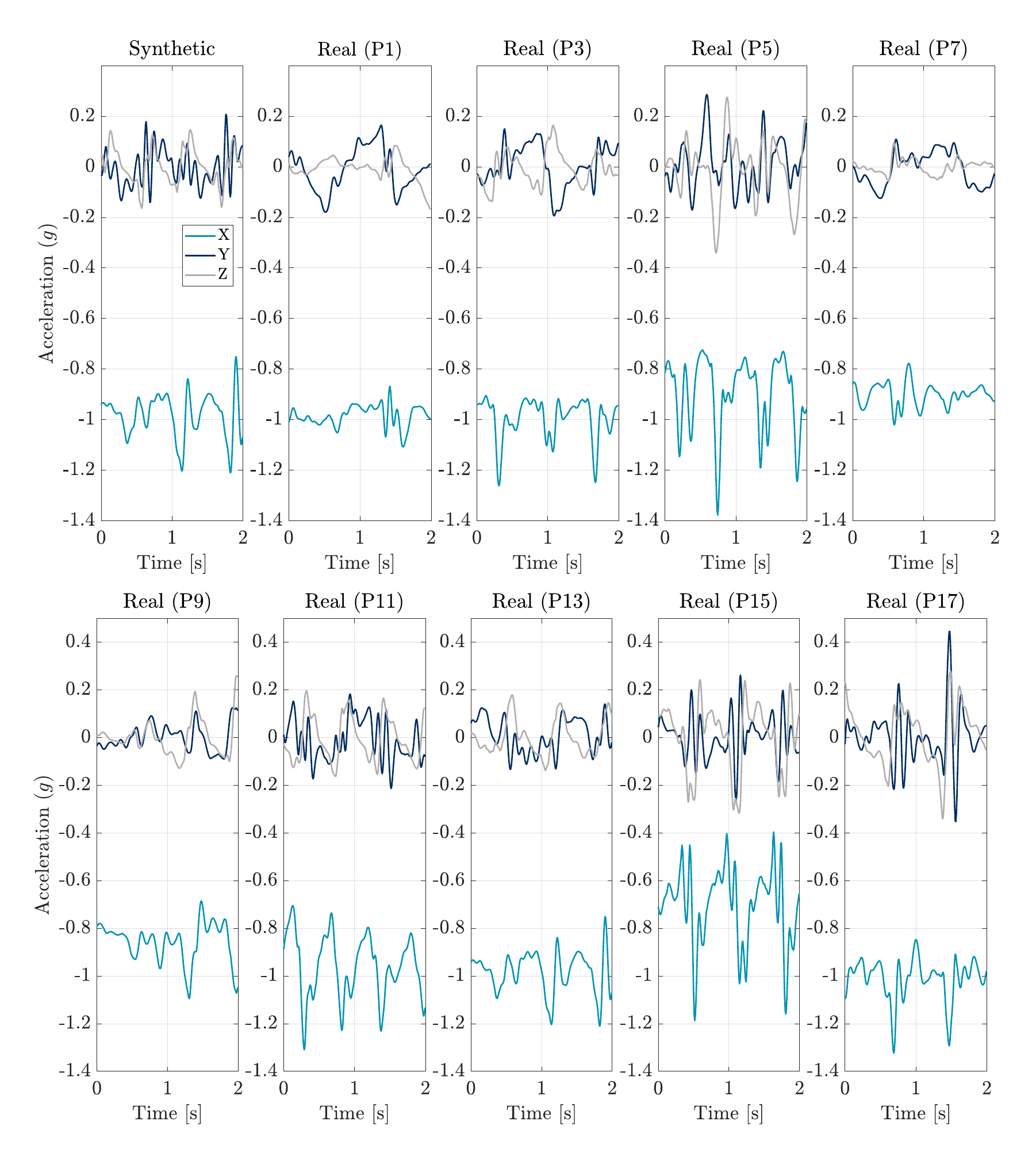}
    \caption{Example tri-axial physical activity signals during walking from synthetic and real datasets. Data from all 24 participants were used to generate the synthetic signals. 9 participants (i.e., P1, P3, P5, P7, P9, P11, P13, P15, and P17) were randomly sampled to provide a general impression of the real signals.}
    \label{fig:DBA_plots}
\end{figure}

Class-conditional value distributions of the selected features for the real and synthetic datasets are shown in Fig.~\ref{fig:feature_boxplots}. In both datasets, the selected features displayed distinct value ranges across activity classes, with mean acceleration features separating static postures and Min-X UT and RMS-X UT differentiating standing from walking. Feature separability appeared stronger in the synthetic dataset, consistent with lower within-class variability. Visual comparison with the representative DBA traces (Fig.~\ref{fig:DBA_plots}) did not suggest systematic discrepancies between the real and synthetic distributions, indicating preservation of the principal activity-related signal characteristics.

\subsection{Model evaluation}
\subsubsection*{Control condition model}
The classification performance of the CCM estimated through leave-one-subject-out cross-validation is shown in Table \ref{tab:CCM_performance}. Overall, the CCM yielded a mean precision of 0.932$\pm$0.045, a mean recall of 0.885$\pm$0.064, and a mean F1-score of 0.896$\pm$0.075 across all 24 participants. Upon examining the recognition rates of the individual activities, inter-person variability in classification performance appeared to be large for recognition of walking (F1-score range: 0.107-1.0), lying down (F1-score range: 0.154-1.0), and transfers (F1-score range: 0.176-0.929). 

% \begin{threeparttable}
\begin{table}[H]
    \centering
    \caption{Performance of control condition model evaluated with leave-one-subject-out cross-validation.}
    \resizebox{\textwidth}{!}{
    \begin{tabular}{l l l l l l l}
    \toprule
         \multicolumn{1}{l}{\textbf{\makecell[tl]{Physical\\activity}}} &
         \multicolumn{2}{l}{\textbf{Precision}} & \multicolumn{2}{l}{\textbf{Recall}} & 
         \multicolumn{2}{l}{\textbf{F1-score}} \\ 
         \cmidrule(lr){2-3}
         \cmidrule(lr){4-5}
         \cmidrule(lr){6-7}
         & Mean $\pm$ SD & Range & Mean $\pm$ SD & Range & Mean $\pm$ SD & Range \\

         \midrule
         Walk & 0.894$\pm$0.072 & 0.744-1.0 & 0.861$\pm$0.197 & 0.058-1.0 & 
         0.861$\pm$0.172 & 0.107-1.0 \\
         
         Stand & 0.914$\pm$0.070 & 0.708-1.0 & 0.912$\pm$0.086 & 0.654-1.0 & 
         0.911$\pm$0.066 & 0.680-1.0 \\
         
         Sit & 0.994$\pm$0.006 & 0.977-1.0 & 
         0.999$\pm$0.002 & 0.993-1.0 & 
         0.997$\pm$0.003 & 0.987-1.0 \\
         
         Lie down & 0.986$\pm$0.037 & 0.833-1.0 & 
         % Recall range: 0.139-1.0
         0.956$\pm$0.187 & 0.083-1.0 & 
         % F1-range: 0.244-1.0
         0.955$\pm$0.172 & 0.154-1.0 \\
         
         Transfer & 0.871$\pm$0.187 & 0.100-1.0 & 0.698$\pm$0.130 & 0.438-0.867 & 
         0.755$\pm$0.159 & 0.176-0.929 \\
         \bottomrule
    \end{tabular}}
    \label{tab:CCM_performance}
\end{table}

\subsubsection*{Feature intervention model}
% \begin{threeparttable}
\begin{table}[H]
    \centering
    \caption{Performance of the feature intervention model evaluated with leave-one-subject-out cross-validation.}
    \resizebox{\textwidth}{!}{
    \begin{tabular}{l l l l l l l}
    \toprule
         \multicolumn{1}{l}{\textbf{\makecell[tl]{Physical\\ activity}}} &
         \multicolumn{2}{l}{\textbf{Precision}} & \multicolumn{2}{l}{\textbf{Recall}} & 
         \multicolumn{2}{l}{\textbf{F1-score}} \\ 
         
         \cmidrule(lr){2-3}
         \cmidrule(lr){4-5}
         \cmidrule(lr){6-7}
         & Mean $\pm$ SD & Range & Mean $\pm$ SD & Range & Mean $\pm$ SD & Range \\

         \midrule
         Walk & 0.922$\pm$0.037 & 0.855-1.0 & 
         0,889$\pm$0.138 & 0.288-1.0 & 
         0.896$\pm$0.100 & 0.448-0.966 \\
         
         Stand & 0.924$\pm$0.051 & 0.768-1.0 & 
         0.932$\pm$0.051 & 0.769-1.0 & 
         0.927$\pm$0.039 & 0.833-0.975 \\
         
         Sit & 0.994$\pm$0.008 & 0.961-1.0 & 
         0.999$\pm$0.001 & 0.996-1.0 & 
         0.997$\pm$0.004 & 0.980-1.0 \\
         
         Lie down & 0.929$\pm$0.205 & 0-1.0 & 
         % Recall range 0.826-1.0
         0.948$\pm$0.204 & 0-1.0 &
         % F1 range 0.857-1.0
         0.937$\pm$0.202 & 0-1.0 \\
         
         Transfer & 0.909$\pm$0.136 & 0.417-1.0 & 
         0.758$\pm$0.151 & 0.500-1.0 & 
         0.816$\pm$0.120 & 0.500-1.0 \\
         \bottomrule
    \end{tabular}}
    \label{tab:FIM_performance}
\end{table}

The classification performance of the FIM estimated through leave-one-subject-out cross-validation is shown in Table \ref{tab:FIM_performance}. Overall, the FIM yielded a mean precision of 0.936$\pm$0.055, a mean recall of 0.905$\pm$0.066, and a mean F1-score of 0.915$\pm$0.064 across all 24 participants. Upon examining the recognition rates of the individual activities, inter-person variability in classification performance appeared to be large for recognition of lying down (F1-score range: 0-1.0), and moderately large for walking (F1-score range: 0.448-0.966) and transfers (F1-score range: 0.500-1.0). 

\subsubsection*{Performance comparison}
On average, the FIM achieved the highest overall F1-score and the lowest inter-participant variability (0.915 ± 0.064), outperforming the CCM (0.896 $\pm$ 0.075). Pairwise comparisons demonstrated a statistically significant difference between models (Friedman test, p-value $<$ 0.05), with post-hoc Wilcoxon signed-rank tests indicating that the FIM significantly outperformed the CCM (p-value $<$ 0.05).

In comparison with the CCM, the FIM provided consistent participant-level improvements, particularly for transfers. The FIM increased the transfer F1-score for 18 of 24 participants, maintained identical performance for one participant, and reduced performance for five. Notably, the per-participant transfer F1-score range increased from 0.176 -- 0.929 under the CCM to 0.500 -- 1.0 under the FIM, indicating a statistically significant upward shift in the lower bound (p-value $<$ 0.05). 

\subsubsection*{Failure cases of the best classification model}
The best classification model, FIM, still produced relatively poor F1-scores ($<$ 0.5) for lying down and walking in two participants. Specifically, all 12 cases of lying down were misclassified as sitting for participant 18, and 33 out of 48 cases of walking were misclassified as standing for participant 1. Through analyzing the gait patterns, it was observed that most participants, including the best classified participant, exhibited Mean-X LB values between -0.1 and 0.1~g. This acceleration range entailed that the vertical axis of the upper body was oriented (nearly) parallel to the bed's surface while lying down. By comparison, for participant 18, the Mean-X LB values were more deviant with fluctuations around -0.4~g, which were in the vicinity of the Mean-X LB range observed during sitting. As for participant 1, based on Min-X UT, the walking accelerations exhibited considerably smaller amplitudes than those of most participants, and the difference is especially noticeable compared to the best classified participant. As shown in Figure \ref{fig:feature_boxplots}, the Min-X UT was found to separate most cases of walking and standing from each other. However, the low amplitudes of participant 1 caused the Min-X UT values during walking to fall within the Min-X UT value range observed during standing. 

\section{Discussions}
Despite the absence of hip fracture patients in our study cohort, the contributions of this work are valuable regardless. Contrary to commercially available activity trackers, we developed our HAR algorithm in a demographic age group (age $\geq$ 80) that is representative of the hip fracture patient population. Beyond demonstrating that continuous physical activity monitoring in older adults can be achieved under simulated free-living conditions and across all clinically relevant functional milestones (walking, standing, sitting, lying, and transfers), this study introduced a structured strategy to manage inter-individual variability using synthetic gait representations. The synthetic data supported the multistep FS pipeline in identifying features that generalized better across different individuals. By using the proposed feature set selected from the synthetic data to train a classifier on the real data, i.e., the FIM, the classification performance and robustness to inter-person variability improved slightly with statistical significance compared to the baseline model CCM. Such improvements were most noticeable in the classifications of transfers. 

\subsection{Real vs. synthetic data}
The differences between the real and synthetic data are worth examining to infer how they complement each other in the development of the FIM. We previously observed that the synthetic data exhibited a lower intra-activity variation than the real data. It is postulated that this was due to a reduced inter-person variability, since DBA produced generalized gait patterns based on the common characteristics across different individuals. This reduction in individual gait deviations can be interpreted as a form of noise suppression, which facilitates the identification of stable and representative activity patterns. This interpretation is consistent with Bai et al.~\cite{bai2022are}, who reported that reduced variability in HAR data improves the stability of pattern recognition in machine learning.

For the final feature set used in the FIM, only the features selected from the synthetic dataset were retained. Although two upper-thigh mean acceleration features (Mean-X UT and Mean-Z UT) overlapped between the real and synthetic selections, one feature selected exclusively from the real data (Min-X LB) was deliberately excluded. Because the synthetic data effectively suppresses individual-specific variability, feature selection performed on the synthetic dataset is more likely to capture fundamental, subject-invariant activity characteristics rather than incidental patterns driven by sampling noise or subject-specific gait traits in the real dataset. Moreover, consistently stronger feature stability and class separability were observed in the synthetic data, while visual inspection of class-conditional distributions and representative DBA traces confirmed that the principal activity-related signal structure was preserved without systematic distortion. Therefore, prioritizing synthetic data-selected features supports improved robustness and generalizability of the FIM, particularly for deployment in heterogeneous clinical populations. Nevertheless, as future work, it is worth investigating the impact of combining features selected from both the real and synthetic data.

\subsection{Synthetic data-selected feature analysis}
The FIM was able to classify most physical activities correctly using solely six features selected from the synthetic data with the multistep FS method. The features included Mean-X UT, Mean-Z UT, Mean-X LB, Min-X UT, RMS-X UT, and RMS-X LB. As shown in Figure~\ref{fig:feature_boxplots}, we observed that the mean accelerations primarily supported the discrimination between standing, sitting, and lying down. These findings are in line with previous HAR studies, which characterized body postures using the mean. For example, Capela et al.~\cite{capela2015feature} tested three different FS algorithms and found that mean accelerations were consistently selected to distinguish between sitting, standing, and lying down, in able-bodied older adults (74$\pm$6.3 years). Similarly, Pannurat et al.~\cite{pannurat2017analysis} found that mean accelerations of the waist were ranked among the six most important features for classification of walking, standing, sitting, and lying down in healthy adults (average age of 67.5 years). Bijnens et al.~\cite{bijnens2019optimization} demonstrated the feasibility of using mean accelerations of the upper thigh to distinguish between standing, sitting, and lying down in healthy older adults between the ages of 60-88 years. Hence, our findings underscore and affirm the importance of the mean acceleration values in distinguishing between static activities whose differences are primarily defined by posture. To distinguish between standing and walking, which are similar in posture, Min-X UT and RMS-X UT appeared to provide discriminative information. Despite the widespread use of the minimum~\cite{chernbumroong2013elderly,erdacs2016integrating,janidarmian2017comprehensive} and RMS~\cite{janidarmian2017comprehensive,bennasar2022significant,chen2021improving} in HAR, their exact discriminative properties are ill-defined in the literature. Nevertheless, it is known that the RMS is sensitive to signal variability and that the minimum is sensitive to signal amplitudes. Both of these properties are distinctive for static and dynamic activities, which could explain their ability to discriminate between standing and walking.

The contribution of the aforementioned features in relation to the recognition of transfers remains challenging to explain. The primary challenge resides in the fact that transfers always occur in between two other activities, which causes their feature values to overlap. In the feature selection study by Capela et al.~\cite{capela2015feature}, no generalizable features for the characterization of transfers could be identified. However, others found that accelerations of the chest, waist, and upper thigh effectively captured the range of motion of transfers, and thereby enhanced recognition rates~\cite{atallah2011sensor}. We extracted features from two similar locations: the lower back and upper thigh. We demonstrated that the six features extracted from these locations could collectively recognize transfers accurately with an F1-score of 0.816$\pm$0.120. Hence, since the most crucial features could not be pinpointed from a univariate standpoint, it is postulated that interactions between these features underpinned the success of the good transfer recognition rate. 

\subsection{Two failure cases}
Although our FIM model generally performed well on most participants, some pronounced misclassification errors were found for participants 18 and 1. First, the FIM model misclassified all instances of lying down as sitting for participant 18. Among participants for whom lying down was mostly correctly classified, the Mean-X LB values primarily varied between -0.1 and 0.1~g. By contrast, the Mean-X LB values of participant 18 were near -0.4~g, which may also be observed during a backward leaning sitting posture. This could potentially explain why instances of lying down were misclassified as sitting for participant 18. There are several external factors that may also have caused this deviating feature value, such as physiological differences between participants, which altered the sensor orientation or displacement during the execution of ADL tasks. Clinical practitioners may need to be particularly mindful of the latter, as displacements are more likely to occur during long-term monitoring in FLC~\cite{atallah2011sensor}, which may harm the HAR algorithm's ability to accurately recognize physical activities~\cite{banos2014dealing}. Second, the FIM model misclassified most instances of walking as standing for participant 1. The misclassifications were postulated to be due to the fact that participant 1 showcased noticeably smaller acceleration amplitudes during walking, which decreased the discriminative properties of Min-X UT to distinguish it from standing. This may explain why walking was often mistaken for standing in participant 1. 

\subsection{Failure cases' implication on real-world patients}
These two failure cases indicate that our HAR algorithm recognizes ambulation behaviors less accurately in individuals whose walking accelerations exhibit lower amplitudes, e.g., due to slower or shuffling gait. Such a limitation presents a potential challenge when applying our proposed HAR algorithm to the hip fracture patient population. Because these forms of ambulation may be more prominent among geriatric hip fracture patients, considering that more than half of them do not regain their prefracture mobility level within the first postoperative year~\cite{bertram2011review,vochteloo2013more}. In addition, following hip fracture surgery, increased double support time, increased single support asymmetry, decreased cadence, and increased step length may be observed among such patients~\cite{gausden2018gait}. These factors highlight the need to consider the unique gait characteristics of hip fracture patients.

Apart from deviating ambulation patterns, transfers may also carry different characteristics in hip fracture patients. Compared to healthy older adults (69.4$\pm$10.9 years), patients recovering from a hip fracture (76.4$\pm$7.1 years) rely significantly more on force compensations from the contralateral side of the fractured hip to perform sit-to-stand transfers~\cite{houck2011analysis}. Besides force asymmetry in the lower extremities, transfer durations are generally prolonged for hip fracture patients. The cohort examined in our study showcased an average sit-to-stand transfer duration of 2.16 seconds. This is comparable to other studies examining healthy older adults, in which an average sit-to-stand transfer duration of 2.41-2.90 seconds was reported~\cite{arcelus2009determination,goffredo2009markerless,salarian2007ambulatory,najafi2002measurement}. However, for rehabilitating hip fracture patients, the average transfer duration was previously estimated at approximately 5.35 seconds~\cite{arcelus2009determination}. Similar deviations in movement symmetry, timing, and motor control are also commonly observed in other neurological and musculoskeletal populations, such as stroke survivors and patients with Parkinson’s disease, suggesting that these generalization challenges may extend beyond hip fracture patients. In conclusion, the deviating transfer movements and prolonged transfer durations highlight the potential generalization problems that may be encountered when using our HAR algorithm in the real-world patient population. 

\subsection{Future work}
Based on the aforementioned limitations and in line with a systematic review on HAR for health research~\cite{straczkiewicz2021systematic}, it is imperative to further develop our HAR algorithm in a diverse participant pool, especially including real-world hip fracture patients, as well as patients that survived a stroke or have Parkinson's, to improve the algorithm's population-level impact. In addition, future work should also explore more expressive synthetic generation schemes beyond DBA. While DBA effectively captures average gait characteristics and suppresses inter-individual variability, it smooths temporal dynamics and may insufficiently represent atypical or pathological patterns such as slow, shuffling, or asymmetric gait in patients with neurological or musculoskeletal conditions. Deep generative models and hybrid artificial intelligence approaches that embed biomechanical or physiological constraints could generate more diverse yet physiologically plausible motion patterns, enabling targeted augmentation of underrepresented behaviors and further improving robustness and generalizability of our HAR algorithm in heterogeneous clinical populations. 

\section{Conclusion}
This study focused on developing a continuous physical activity monitoring system that leverages a novel synthetic data generation scheme to aid in the multistep feature selection process, thereby enhancing HAR performance and generalizability across individuals. The cohort consisted of healthy older adults aged $\geq$ 80 years, providing an age-matched first step toward future deployment in hip fracture rehabilitation. The investigated activities comprised walking, standing, sitting, lying down, and postural transfers. First, intra-activity variation was addressed by collecting sensor data under simulated free-living conditions. Second, robustness to inter-person variability was supported by generating a synthetic data set that represented common gait characteristics across individuals. We found that feature selection based on synthetic data improved downstream classification on real data, with the most notable gains observed in the recognition of transfers. The resulting monitoring system demonstrated good predictive performance using only six features, selected based on synthetic data and derived from two accelerometers placed on the upper thigh and lower back. While these preliminary results are encouraging, further validation in real-world patient populations remains essential to determine the clinical utility of the proposed monitoring system.

\begin{credits}
\subsubsection{\ackname}We acknowledge Michael Bui for contributions to the initial algorithm implementation and data post-processing pipelines supporting this work

This research work is funded by HealthyW8 and Up\&Go projects.

On behalf of the Up\&Go after a hip fracture group:
B. Broersma, K. Brouwer, E. C. Folbert, T. Gerrits, S. M. Gommers, A. J. M. Harperink, P. T. Hofstra, M. M. Kemerink op Schiphorst, N. M. Lammerink-Smienk, M. P. Luttje, D. K. Marissen-Heuver, P. M. M. Mars, M. A. H. Nijhuis-Geerdink, W. S. Nijmeijer, A. H. S. Oude Luttikhuis, T. M. Oude Weernink, C. de Pagter, J. Schokker-Viergever, R. T. J. Vlaskamp, M. Voortman, and S. Woudsma

On behalf of the HealthyW8 working group:
- LIH: Torsten Bohn, Farhad Vahid, Laurent Malisoux, Yvan Devaux, Manon Gantenbein, Michel Vaillant, Jonathan Turner, Mahesh Desai, Adriana Voicu, Alejandra Loyola Leyva
- LIST: Christoph Stahl, Yannick Naudet
- NIUM: Alberto Noronha, Adam Selamnia, Jorge Ribeiro, Fernando Veloso
- DFKI: Serge Autexier, Anke Königschulte
- Virtech: Roumen Nikolov, Vladislav Jivkov, Alexandre Chikalanov
- BIPS: Sarah Forberger
- SPORA: Mireia Faucha, Hernández, Sergio Yanes Torrado, Mariona Estrada Canal
- CREDA: Zein Kallas, Amelia Sarroca, Djamel Rahmani
- USG: Maria Giovanna Onorati, Gino Gabriel Bonetti
- CNR: Arianna D`Ulizia, Alessia D`Andrea
- CITA: Tiziana De Magistris
- UEV: Elsa Sousa De Lamy
- IDISBA: Josep A. Tur, Cristina Bouzas
- AOUBO: Giuseppe Tarantino, Michele Stecchi, Lucia Brodosi
- DTU: Rikke Andersen, Gitte Ravn-Haren
- UT: Ying Wang, Shuhao Que
- UC: Daniela Rodrigues
- RCNE: Yoanna Ivanova, Boyko Doychinov
- TU/e: Pieter Van Gorp, Astrid Kemperman
- MEDEA: Pietro Dionisio, Francesco Agnoloni
- EADS: George-Mihael Manea
- ENHA: Joost Wesseling, Konstantina Togka
- KNEIA: Ciro Avolio, Cristina Barragan Yebra, Christina Barragan Mesa
- EFAD: Marianna Kalliostra, Ezgi Kolay, Katarzyna Janiszewska”
\end{credits}
\bibliographystyle{splncs04}
\bibliography{mybibliography}

\end{document}